Research Paper

# From slides to AI-ready maps: Standardized multi-layer tissue maps as metadata for artificial intelligence in digital pathology

Gernot Fiala [a,1], Markus Plass [a,1], Robert Harb [a,b], Peter Regitnig [a], Kristijan Skok [a,c], Wael Al Zoughbi [a], Carmen Zerner [a], Paul Torke [a], Michaela Kargl [a], Heimo Müller [a,h,*], Tomas Brazdil [d], Matej Gallo [d], Jaroslav Kubín [d], Roman Stoklasa [d], Rudolf Nenutil [e], Norman Zerbe [i,j,k,l], Andreas Holzinger [a,f], Petr Holub [g,h]

[a] Medical University of Graz, Diagnostic and Research Institute of Pathology, Graz, Austria
[b] Graz University of Technology, Institute of Computer Graphics and Vision, Graz, Austria
[c] University of Maribor, Institute of Biomedical Sciences, Maribor, Slovenia
[d] Masaryk University, Faculty of Informatics, Brno, Czech Republic
[e] Masaryk Memorial Cancer Institute, Department of Pathology, Brno, Czech Republic
[f] University of Natural Resources and Life Sciences Vienna, Human-Centered AI Lab, Institute of Forest Engineering, Department for Ecosystem Management, Climate and Biodiversity, Vienna, Austria
[g] Masaryk University, Institute of Computer Science, Brno, Czech Republic
[h] BBMRI-ERIC, Graz, Austria
[i] Institute of Pathology, University Hospital RWTH Aachen, Aachen, Germany
[j] Charité - Universitätsmedizin Berlin, Corporate Member of Freie Universität Berlin and Humboldt-Universität zu Berlin, Institute of Medical Informatics, Berlin, Germany
[k] EMPAIA International e.V., Berlin, Germany
[l] European Society of Digital and Integrative Pathology, Lisbon, Portugal

## ARTICLE INFO

## ABSTRACT

A Whole Slide Image (WSI) is a high-resolution digital image created by scanning an entire glass slide containing a biological specimen, such as tissue sections or cell samples, at multiple magnifications. These images are digitally viewable, analyzable, and shareable, and are widely used for Artificial Intelligence (AI) algorithm development. WSIs play an important role in pathology for disease diagnosis and oncology for cancer research, but are also applied in neurology, veterinary medicine, hematology, microbiology, dermatology, pharmacology, toxicology, immunology, and forensic science.

When assembling cohorts for AI training or validation, it is essential to know the content of a WSI. However, no standard currently exists for this metadata, and such a selection has largely relied on manual inspection, which is not suitable for large collections with millions of objects.

We propose a general framework to generate 2D index maps (tissue maps) that describe the morphological content of WSIs using common syntax and semantics to achieve interoperability between catalogs. The tissue maps are structured in three layers: source, tissue type, and pathological alterations. Each layer assigns WSI segments to specific classes, providing AI-ready metadata.

We demonstrate the advantages of this standard by applying AI-based metadata extraction from WSIs to generate tissue maps and integrating them into a WSI archive. This integration enhances search capabilities within WSI archives, thereby facilitating the accelerated assembly of high-quality, balanced, and more targeted datasets for AI training, validation, and cancer research.

## 1. Introduction

A WSI is a high-resolution digital image that is created by scanning an entire glass slide containing tissue sections or cell samples of a biological specimen at high magnification. These gigapixel images, with resolutions of up to $150,000 \times 150,000$ pixels and pixel sizes of down to 0.125 μm are represented as image pyramids with different magnification levels, shown in Figs. 1(a) and 1(b) based on an example from The Cancer Genome Atlas (TCGA) [1,2].

* Corresponding author at: Medical University of Graz, Diagnostic and Research Institute of Pathology, Graz, Austria.
*E-mail address:* heimo.mueller@medunigraz.at (H. Müller).
[1] These authors contributed equally to this work.

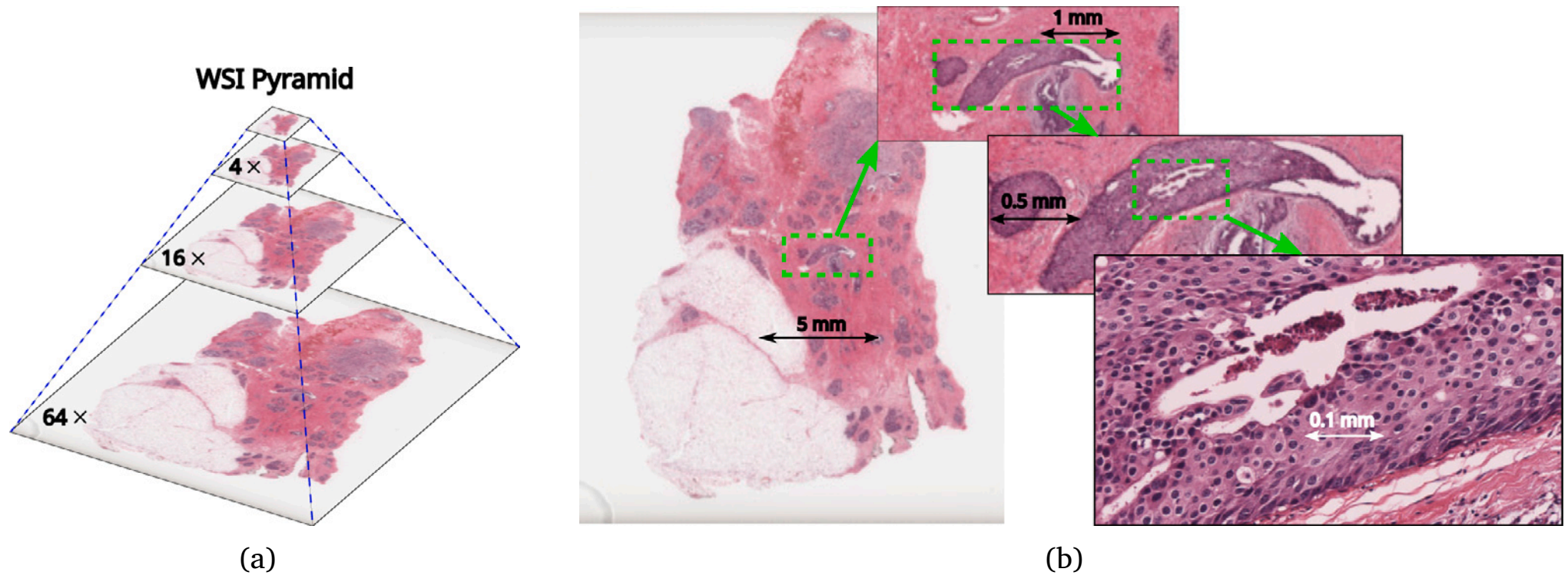

**Fig. 1.** WSI example from TCGA breast dataset. (a) WSI pyramid with down-sampled pyramid levels. (b) Details at different magnification levels.

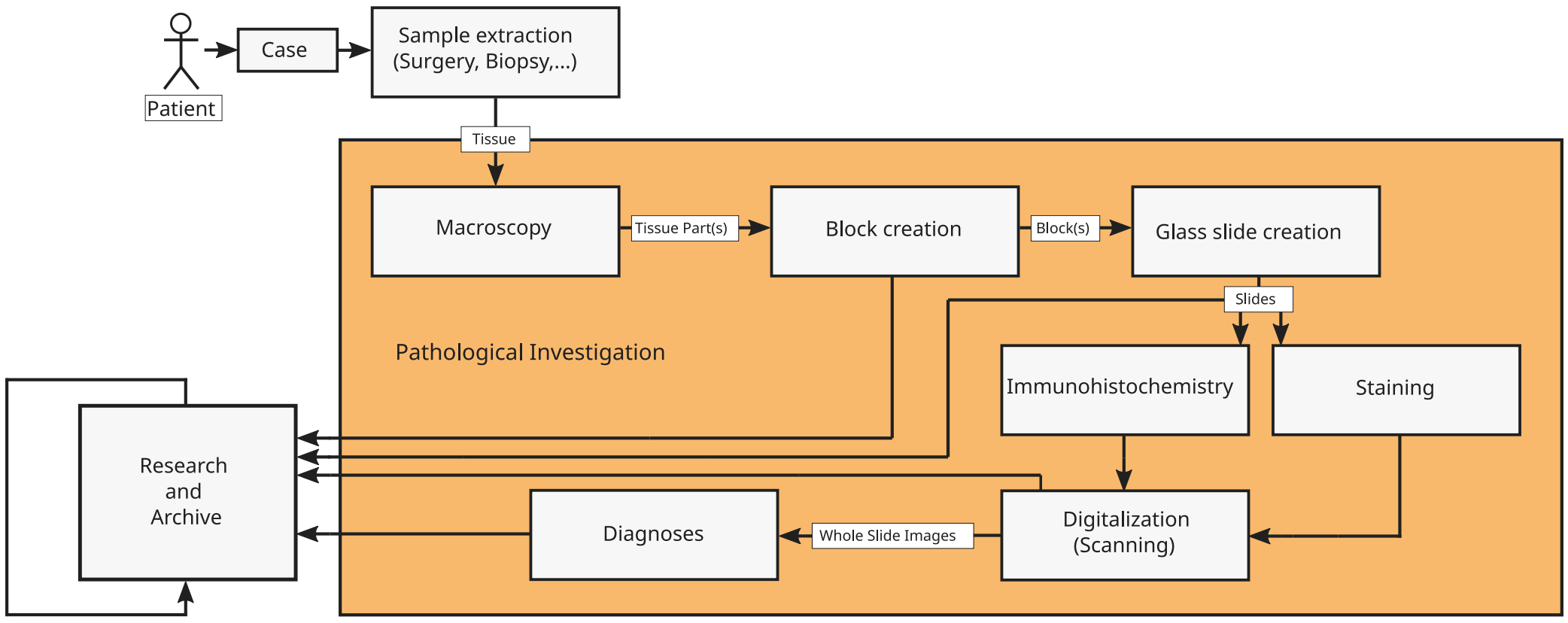

**Fig. 2.** Overview of a pathological workflow.

Such images are viewable, analyzable, and shareable digitally, and are increasingly used to develop AI algorithms. There are various applications of WSIs in fields such as pathology where they are gaining increasing relevance for diagnosing diseases and oncology where they play a big role in cancer research [3–5].

Other applications include neurology and veterinary medicine [6, 7], as well as hematology, microbiology, dermatology, pharmacology, and toxicology [8–11], with further use in immunology and forensic science.

The process of a typical pathological case is shown in Fig. 2. Tissue samples are obtained from patients via biopsy or surgery, processed into blocks, sectioned, mounted on glass slides, and stained using hematoxylin and eosin (H&E), special stains, or immunohistochemistry (IHC). Then, the slides are digitized by scanners and provided for diagnostics. Finally, the diagnosis, scanned slides (WSIs), blocks and glass slides are archived. Pathologists and researchers can request slides from the archive for research purposes. A case can contain multiple staining or IHC steps, resulting in several WSIs per case.

Understanding the specimen composition of a WSI is crucial when creating cohorts for research and especially for training or validating AI algorithms. The increasing digitization of pathology archives highlights the need for automated AI-driven metadata extraction and the standardization of metadata in WSI archives, as demonstrated by pathology data from the Medical University of Graz.

Between 1984 and 2019, pathologists at the Medical University of Graz examined 7,430,001 slides from 1,040,984 cases. That makes around 7 slides per case across different pathological diagnoses.

In the year 2023 there were 54,558 total cases with 397,942 slides at the Medical University of Graz. Some of the cases are listed in Table 1 with additional information of the cases. There were 7830 colon cases with 31,920 slides. This results in around 4 slides per colon case. Out of the total slides, 876 contained lymph nodes and 2225 slides were treated with IHC. Numbers for the organs prostate, mamma, lung, liver, kidney, pancreas and stomach are also presented. For pancreas, the average number of slides is almost 18. In some cases there are fewer slides and sometimes there are up to 50 slides per case or even more. This shows the complexity of the pathological work. Information as shown in Table 1 can be extracted from pathology laboratory information and management systems (LIMS).

**Table 1**
Statistics of pathological cases for selected organs at Medical University of Graz from the year 2023.

| Organ | # Cases | Slides | Av. Slides/Case | Lymph | IHC |
|---|---|---|---|---|---|
| Colon | 7830 | 31 920 | 4.07 | 876 | 2225 |
| Prostate | 1989 | 48 398 | 7.27 | 46 | 782 |
| Mamma | 1461 | 14 464 | 10.99 | 86 | 3385 |
| Lung | 1058 | 10 090 | 9.54 | 396 | 1808 |
| Liver | 585 | 6102 | 10.43 | 17 | 1612 |
| Kidney | 583 | 9396 | 16.12 | 42 | 2184 |
| Pancreas | 333 | 5831 | 17.51 | 314 | 575 |
| Stomach | 149 | 1630 | 10.94 | 291 | 158 |

Pathologists and researchers often request WSIs from biobanks, such as the Biobank Graz of Medical University of Graz [12], to compare similar cases or train AI algorithms to assist in the diagnosis of cancer [13–16], segment tissue types [17–20] or make survival predictions [2,21,22]. Currently, WSIs are provided as whole cases rather than individual slides with specific content, because metadata such as tumor size, tissue types such as fat, nerve, muscle etc., or pathological alterations, such as inflammation, neoplasm or necrosis, are often missing, making the specific contents of slides unknown. Manual microscopic inspection remains the primary method of identifying slide content, but this process is inefficient for large archives.

To support cancer research, including clinical studies and AI development, datasets based on diagnoses, organ types, tissue types, or pathological alterations are required. Standardized metadata enables the creation of WSI catalogs that comply with FAIR principles (Findable, Accessible, Interoperable, and Re-usable) [23], facilitating efficient data management and access. A common metadata format describing the source of the specimen, occurring tissue types, and pathological alterations enables interoperability between slide catalogs, improves reproducibility, data quality, and cohort selection, facilitating large-scale experimentation, reducing bias, and enhancing the generalizability of AI models. Large, annotated datasets are particularly important for training deep learning models [13,16,18,20]. While public datasets such as TCGA [1,2], CAMELYON16 [24,25] and CAMELYON17 [26–28] are available, leveraging biobank collections allows AI algorithms to be trained on hundreds of thousands of WSIs [16,29], providing larger-scale and more clinically relevant datasets [3,4,12]. Yet, AI-driven metadata extraction and the extension of standardized structured AI-ready metadata into WSI archives is still missing.

This paper proposes a metadata model for integrating AI-ready metadata into WSI archives advancing digital pathology by:

- A metadata model with common syntax and semantics to generate multi-layer tissue maps for WSIs and a profiling mechanism for specific application domains. The model describes each WSI of a collection with a low-resolution multi-layer “tissue map” that provides a structured description of each histopathological slide. The tissue map is organized into three layers: Layer 1 captures information about the source material (e.g., the organ), Layer 2 describes common tissue types present, and Layer 3 details any (pathological) alterations observed. The tissue map serves as AI-ready metadata.
- The metadata model contributes to ongoing initiatives to establish community-wide metadata standards, as endorsed by BBMRI-ERIC.
- WSIDOM framework is provided to visualize tissue maps as segmentation maps for human readability.
- A proof-of-concept implementation with binary classifiers demonstrates the automated extraction of WSI metadata to generate tissue maps and shows the effectiveness of the proposed metadata standard for describing WSI content in a structured and interoperable format.
- It showcases the integration of tissue maps into a WSI archive illustrating how this metadata model enhances the organization, searchability, and retrieval of slides, enabling more efficient and targeted access to relevant WSIs. In addition, it supports pathologists and researchers in creating datasets more quickly, while also facilitating the development of more balanced and higher-quality datasets, ultimately benefiting the development of AI models and advancing cancer research.
- Providing Annotations based on the proposed metadata model for the used TCGA dataset.

This paper is organized as follows: Section 2 lists related work in that research field. Section 3 explains the methodology of the proposed metadata standard with tissue maps and its integration into WSI archives, applications and use cases. Section 4 presents the method to extract metadata from WSIs. Section 5 presents the generation of tissue maps based on the extracted data and shows the evaluation of the metadata model standard integrated into a WSI archive. Section 6 discusses different application domains for the metadata model standard and Section 7 concludes the paper and gives an outlook on future work.

## 2. Related work

WSIs are increasingly managed by biobanks [3,4,12] and are accessible either directly or via research infrastructures such as BBMRI-ERIC [30]. Given the sheer volume of WSIs stored across diverse cases in these biorepositories, advanced data access mechanisms are essential. In particular, robust search capabilities are required to identify WSIs corresponding to specific diagnoses or cases, facilitating the construction of curated datasets for tasks such as cancer diagnoses [13–16], tissue segmentation [17–20] or make survival predictions [2,21,22]. Due to the manual effort involved in WSI selection, researchers increasingly explore algorithmic solutions to optimize the search process.

In 2019 Hedge et al. introduced a similar image search for histopathology (SMILY) [31], a deep learning-based framework to search for images in the field of histopathology. Using a selected image patch from a WSI as a query, SMILY identifies similar histological patterns, organ sites, and cancer grades. The underlying neural network was trained on datasets containing general image content (e.g., animals, objects, human faces), eliminating the need for annotated histopathology images. The framework was evaluated on WSIs from prostate, breast, and colon cancers, demonstrating retrieval of relevant histological features including arteries, epithelium, lymphatic vessels, and stroma.

In 2020 Kalra et al. presented Yottixel [32] an image search engine for histopathological WSI. Also based on query image inputs, Yottixel enables high-speed search through large-scale WSI archives, returning similar cases and their diagnostic reports. It was evaluated using over 2000 WSIs from the TCGA dataset for different cancer types and grades.

Building on this, Kalra et al. introduced an advanced AI-based search algorithm [13] further scaling the image-query approach. The model was evaluated with almost 11,000 patient WSIs and over 20 million image patches with 32 cancer types and 25 anatomic sites. Results showed strong performance in retrieving visually and clinically similar images based on patch-level similarity.

Chen et al. used Yottixel [32] as a foundation and extended it to self-supervised image search for histology (SISH) [33]. SISH was trained using slide-level annotations and evaluated across 22,000 cases involving 56 disease types. It demonstrated robustness even in identifying rare cancer subtypes, addressing limitations of data scarcity in conventional models.

While all of these approaches offer powerful patch-level retrieval based on visual similarity, they typically require manual query selection and do not support metadata-based search across entire WSI archives. In contrast, the metadata model with multi-layer tissue maps proposed in this work enables comprehensive search and filtering of WSIs based on structured metadata descriptors.

Metadata extraction from WSIs has also been explored through ML. Weng et al. [34] proposed a multimodal multitask learning framework to predict major slide-level metadata, including tissue type, fixation method, sampling procedure, and staining technique. Their model was trained using information at the patch, slide, and case levels and evaluated on TCGA and internal datasets. However, it lacked granularity regarding tissue subtypes or pathological alterations.

Sirinukunwattana et al. [35] developed a method for WSI segmentation into tissue regions using phenotypic profiling. By classifying cells into four types (necrotic debris, malignant epithelial, inflammatory, and spindle-shaped), and constructing cell interaction networks, they labeled regions with six distinct tissue phenotypes. These metadata were used to predict the risk of distant metastasis in colorectal cancer, though not applied to search or catalog enrichment.

Another example is MIKAIA [17], a histopathology-oriented framework designed for clinical research. It supports WSIs from various scanners and utilizes data augmentation to enable generalization. Trained on slides from six different scanners, MIKAIA can automatically annotate nine tissue and alteration classes with high accuracy [19], and export results in multiple formats. However, while capable of annotating structures such as fat and muscle tissue, its metadata coverage remains limited relative to the needs of large-scale WSI cataloging.

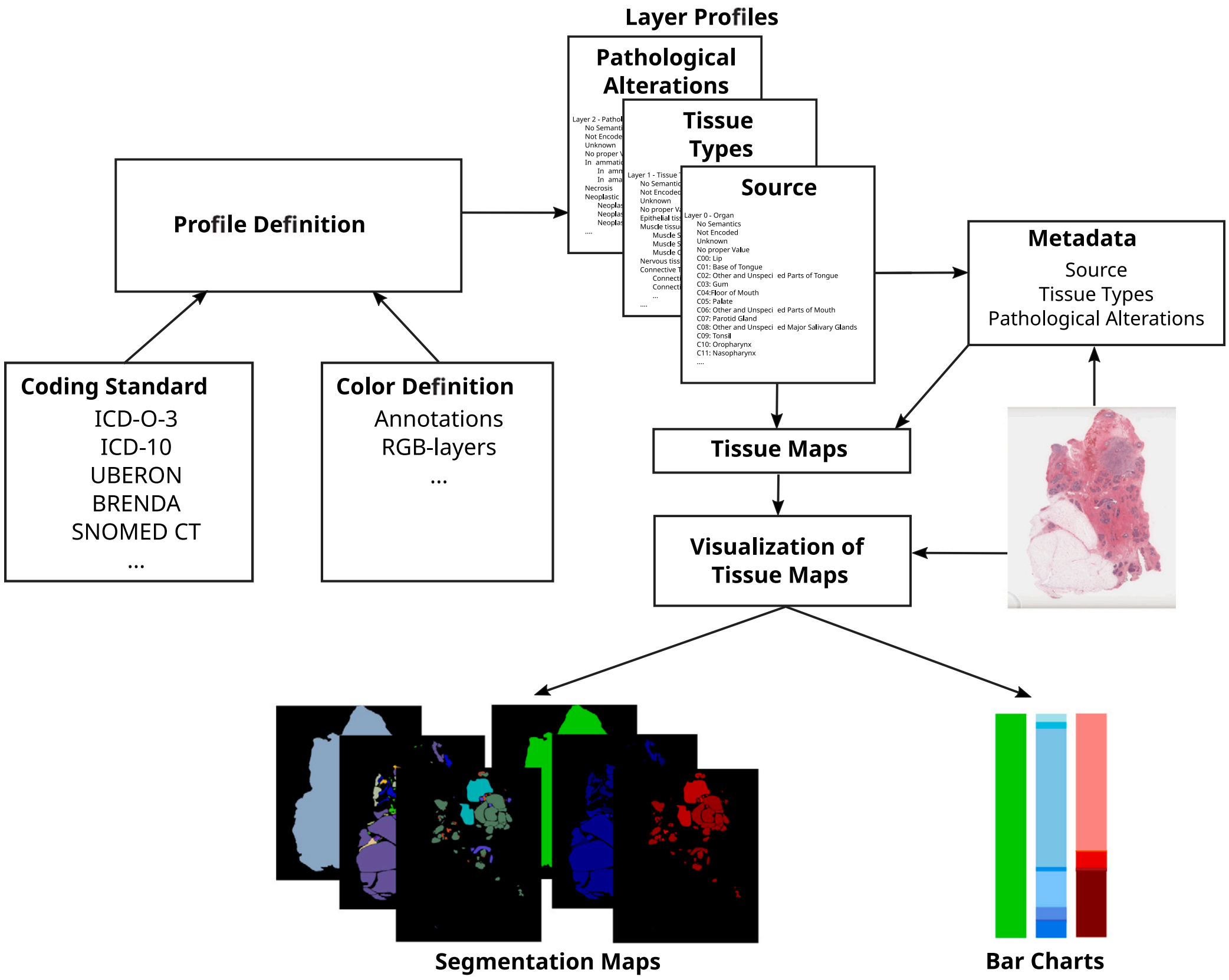

**Fig. 3.** Definition of layer profiles based on coding standards and look-up table to color definitions. The layer profiles are used to define the tissue map content, which can be visualized with different methods. Examples of tissue map visualizations are segmentation maps or bar charts for each WSI.

Collectively, these methods either extract specific metadata or implement image-query search approaches. However, a standardized, comprehensive metadata framework to describe the full content of WSIs across large archives is still lacking. Manual selection of WSIs remains time-consuming, and current methods often do not scale well for large WSI collections. The proposed metadata model, featuring structured multi-layer tissue maps, addresses these limitations. By enabling fine-grained filtering of WSIs based on source (organ), tissue type, and pathological alterations, the model supports the creation of tailored data cohorts, accelerating AI development in digital pathology. The tissue maps are designed to generate AI-ready metadata by capturing these key attributes, ensuring seamless integration with machine learning workflows.

## 3. Proposed model for AI-ready metadata

The domain field using WSIs is characterized by a high degree of heterogeneity, which cannot be sufficiently represented by a simple metadata standard. In addition, general slide-level descriptions are often inadequate, as the spatial distribution of different structures contains important information.

To meet these requirements, we have (a) developed a common denominator for all the different areas in the form of a three-layer model and (b) designed the metadata description as an AI-ready mini-map that encodes spatial information in these three layers.

The three layers, also called "tissue map" defined by the slide metadata model include: (1) the source of the material, (2) the tissue type or other classifications at the microscopic level, and (3) (pathological) alterations or a highlighting layer.

A profile defines the possible values that can be encoded for each layer, as well as their associated semantics through a link to a known ontology. Such encodings in the field of surgical pathology include Uber Anatomy oncology (UBERON) [36,37], Systematized Nomenclature of Medicine Clinical Terms (SNOMED CT) [38], Open Biological Ontologies (OBO) [39], BRaunschweig ENzyme DAtabase (BRENDA) [40,41], International Classification of Diseases of oncology Third Edition (ICD-O-3) [42,43], International Statistical Classification of Diseases and Related Health Problem (ICD-10) [44] or National Cancer Institute Thesaurus (NCIt) [45], shown in Fig. 3.

If a profile is limited to 256 different values across the three layers, it is practically encoded in an 8-bit image and stored in common, naturally lossless compressed image formats. The profile maps each entry to a color code via look-up tables (LUT). This allows the visualization of defined source materials, tissue types and (pathological) alterations as segmentation maps, bar charts or with other methods. In this context, the lossless compressed image format also serves as AI-ready metadata, ensuring both efficient storage and direct usability for downstream analysis.

In the following, we illustrate how we defined the three layer profiles for "surgical pathology". The defined profiles can be found in the Github repository: https://github.com/human-centered-ai-lab/WSIDOM.

### 3.1. Profile definition for tissue maps

The metadata criteria are selected on oncology and vocabulary to make the standard concrete, simple and practical for comprehensive usage. International coding standards are used. In this work, the three layer profiles (source, tissue types, pathological alterations) are based on international coding standards describing organs, tissue types and pathological alterations. The first 4 entries of each profile are defined by NULL values similar to the HL7 standard [46] and describe NI (No Semantics), UNC (Not Encoded), UNK (Unknown) and NV (No proper Value). The rest of the entries are based on the selected coding standards. The source profile is based on ICD-O-3 [42,43] topology, the tissue type profile and the pathological alteration profile use NCIt [45] encoding.

**Table 2**
The breast organ as an example entry of the source material profile.

| Header | Description | Example |
|---|---|---|
| ID | Identifier for entry | 5 |
| PARENT | Link to parent entry ID | −1 |
| CODE | Coding standard | C50 |
| DEF COLOR | Color of annotation | #000037 |
| COLOR | Color for WSIDOM visualization | #000037 |
| NAME | Name of coding standard entry | Breast |
| COMMENT | Additional information | Breast source material |
| ONTOLOGY | URL to coding standard description | https://bioportal.bioontology.org/ontologies/ICD-O-3/?p=classes&conceptid=https%3A%2F%2Fdata.jrc.ec.europa.eu%2Fcollection%2FICDO3_O%23C50 |
| CONCEPT | URL to coding standard entry ID | https://data.jrc.ec.europa.eu/collection/ICDO3_O#C50 |

A structure is defined to describe each entry of the profile, as shown in an example in Table 2. Each entry starts with an ID element as a unique identifier, followed by a PARENT element. If the entry has no parent, a value of '−1' is assigned, otherwise the ID from the parent entry is used. This allows the representation of hierarchies inside the profile and direct integration of international coding standards into the profiles. The CODE element refers to the used code of a coding standard and defines the name of annotations for the selected tissue map. The element DEF COLOR shows the used color in annotation tools like QuPath [47–49], Orbit Image Analysis [50], or V7 Darwin [51]. The COLOR element defines the visualization color for the generated tissue maps. The color values must be encoded in hexadecimal (HEX) format. The NAME element describes the coding and the COMMENT element is used for additional information. ONTOLOGY is an element for a general description or a link to the coding standard description. The CONCEPT element links the coding standard with a concept ID from the applied coding standard, which may reference a website or a document providing a detailed description of the coding. The profiles are stored in csv files. This enables efficient handling of the profiles to generate the AI-ready tissue maps and is easy to interpret for humans. Interpretability, traceability and thus explainability are therefore not only becoming more important, but are actually mandatory, because upcoming regulations and requirements demand explainability [52–54].

**Layer 1** defines the source material of the specimen, namely the organ or organ part type on a WSI. In this work, organ types are mapped to codes based on the top-level oncology classes of the ICD-O-3 standard [42,43]. This standard allows coding for a broader spectrum of specimen material from humans like colon, nose, prostate etc. In total 80 different organ classes are defined at the top-level oncology. Examples of ICD-O-3 codes are C18 for colon, C50 for breast, or C64 for the kidney. Information on organ types may be obtained from medical reports or clinical data, or predicted using ML algorithms. Layer 1 may also serve to define the type of procedure used to obtain the specimen, which can be coded according to ICD-10-PCS [55] or other relevant procedural coding systems.

**Layer 2** defines the tissue types of the specimen. In this work, the layer 2 profile is for standard H&E staining based on the NCIt [45] encoding for different muscle, nerve, connective-tissue types, epithelium and additional tissue categories. These tissue types are structured hierarchically, with general classes at higher levels followed by more detailed subtypes. The profile represents a selected set of tissue types and their hierarchical relationships. Control tissue, which is typically used in the context of immunohistochemistry or special stains, but not in routine H&E staining, may also be considered within this layer. Since tissue type classification in routine diagnostics is primarily based on H&E-stained slides, control tissue generally does not appear in this setting. If classification of slides containing control tissue is required, reliable automated identification would necessitate highly standardized and reproducible placement of the control tissue on the slide, for example in a consistent location opposite the label. Furthermore, the corresponding staining type would need to be encoded in the laboratory information system, and control tissue could be excluded by comparison with the H&E-derived specimen contours to ensure unambiguous technical attribution.

**Layer 3** defines the (pathological) alterations of the specimen. The used profile is based on NCIt [45] coding standard and defines different pathological alterations like inflammation, necrosis, neoplastic-malignant, neoplastic-benign, neoplastic-metastatic etc. A hierarchical structure is used for general and detailed alterations.

Algorithms process the tissue maps, or they are visualized in various ways for human inspection, using different colors or patterns for each layer. In this work, we use LUT for the visualization of the tissue maps. The LUT is used to map the entries of each layer to defined color values to create segmentation maps, shown in Fig. 3. Integrating the established standard into WSI archives supports filtering and faster assembly of high-quality, balanced datasets for research and AI development, while providing AI-ready metadata.

### 3.2. *Tissue map visualization framework*

This work introduces the framework Whole-Slide Image Description of Morphology (WSIDOM) to visualize the content of WSIs based on the defined profiles of the tissue map. WSIDOM uses the metadata included in the three layer-profiles (source material, tissue type, and (pathological) alterations) to create segmentation maps for human readability. The code is located in the Github repository: https://github.com/human-centered-ai-lab/WSIDOM.

An overview of how the framework works is shown in Fig. 4. The framework requires a WSI, the defined layer profiles and metadata for each layer. The profiles of source material, tissue types and pathological alterations provide the LUT to map the entries of each profile to a defined color value. Based on the inputs (i.e., metadata and layer profiles) WSIDOM creates the segmentation maps for source material, tissue types and pathological alterations. The generated segmentation maps have thumbnail sizes with a resolution between 1000 × 1000 and 2000 × 2000 pixels. WSIDOM allows to overlay the WSI input with the segmentation maps, by defining a transparency value alpha. The value of alpha can be set between 0 and 1, where 0 is completely transparent and 1 is not transparent.

An example of the results of WSIDOM for visualizing tissue maps is shown in Fig. 5. The WSI used for this visualization example is part of the breast TCGA dataset. In this example, the metadata in the form of annotations were made with QuPath [47] and exported as geojson files for each layer. The upper row of images in Fig. 5 shows the visualized tissue maps with alpha value set to 1 and different LUT for color visualization. The three layers on the right are represented by single-channel colors and visualized as green, blue and red to store the tissue maps in common, naturally lossless compressed image formats. On the left side, the visualization of the tissue maps is done with annotation colors to be human-readable. In the second row of images in Fig. 5, the alpha value is set to 0.5 to overlay the segmentation maps with the input WSI. The different colors show different classes in the corresponding layer. In the source material layer, only the breast organ type is present. In layer 2, Connective-Tissue (CT), Connective-Tissue-Dense (CT-Dense), Connective-Tissue-Loose(CT-Loose), Connective-Tissue-Blood (CT-Blood), Connective-Tissue-Lymphatic (CT-Lymphatic), Connective-Tissue-Fat (Fat), Nerve and Epithelium are present. Layer 3 shows Neoplastic, Neoplastic-Malignant, Inflammation-Chronic and Not Encoded (UNC) regions.

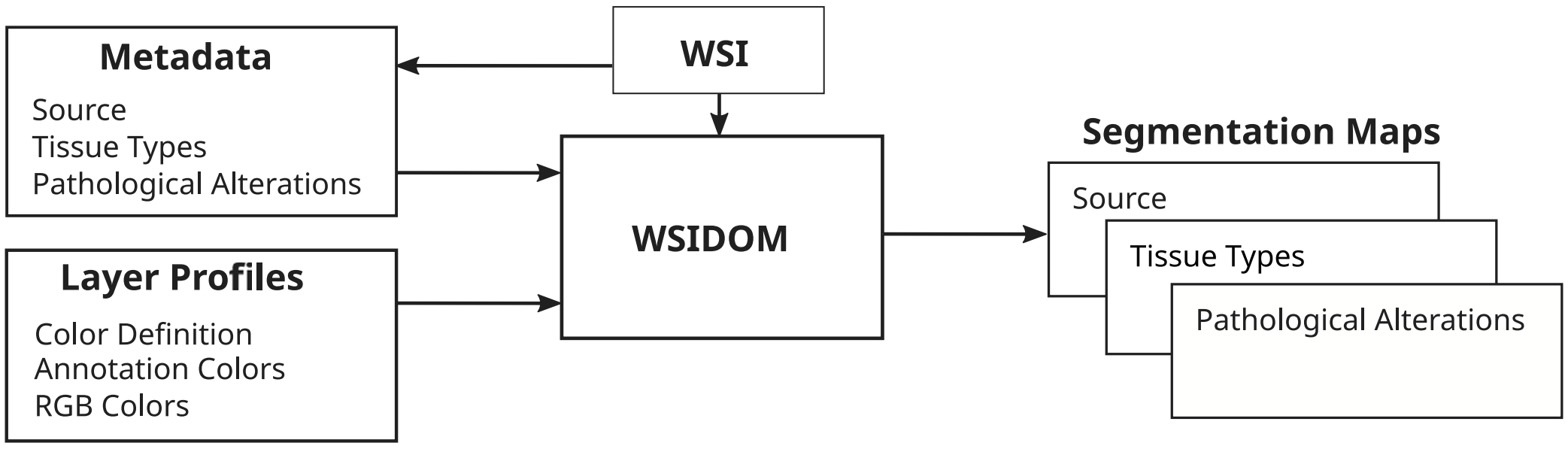

**Fig. 4.** General process to visualize the tissue maps with WSIDOM as segmentation maps.

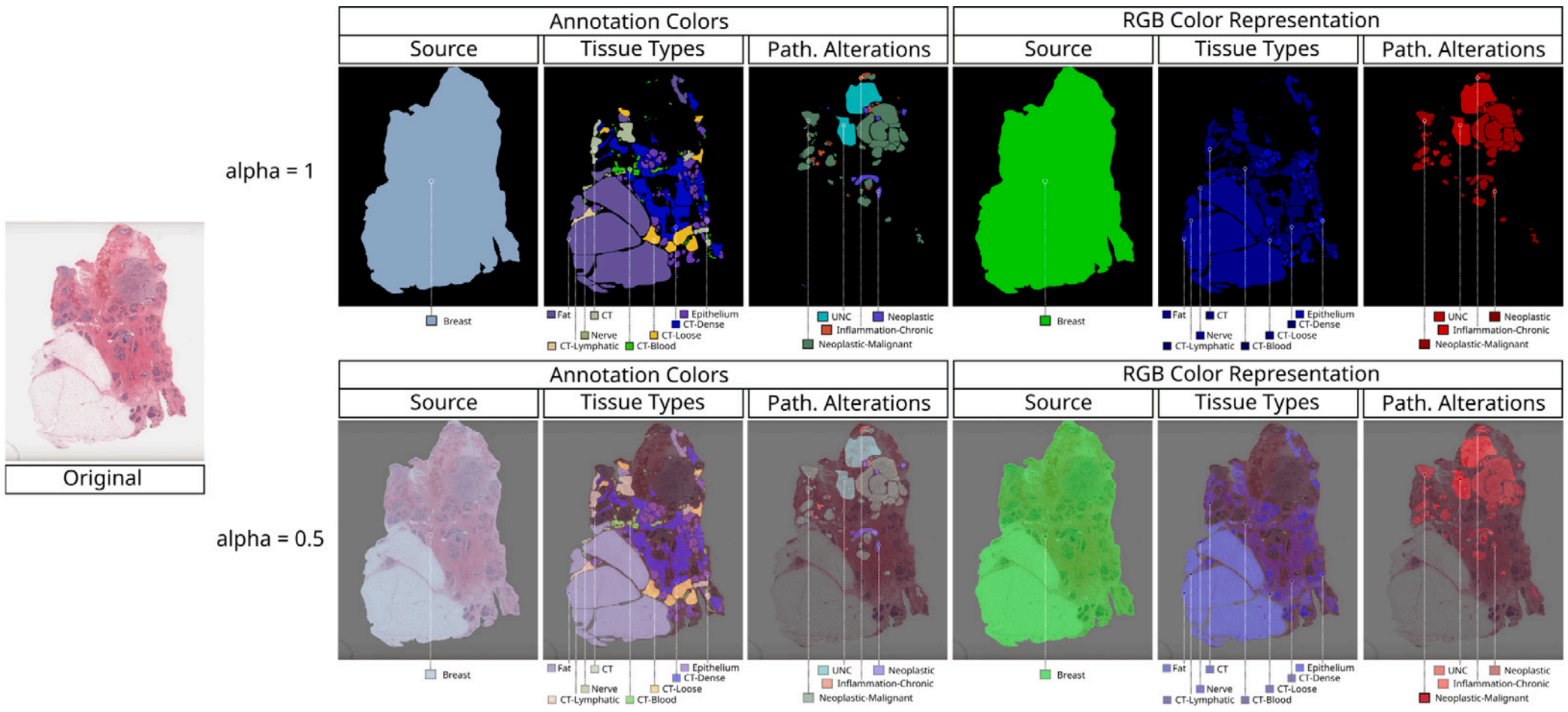

**Fig. 5.** Visualization of tissue maps as segmentation maps with different colors and alpha values. The original WSI from the TCGA dataset is shown on the left side. The upper row shows the segmentation maps with different visualization colors with an alpha value of 1. The lower row shows the segmentation maps overlaid with the WSI based on the same visualization colors and an alpha value of 0.5. (For interpretation of the references to color in this figure legend, the reader is referred to the web version of this article.)

The tissue maps enables the visualization of the metadata in different forms. One way is to use the segmentation maps generated with the WSIDOM framework, as shown in Fig. 5. Another method is to calculate the class areas for each layer and present the normalized ratios as bar charts. One way is to normalize the ratios per resolution of the WSI, other ways are to normalize per specimen area or content area of each layer. Pathologists prefer the normalization per specimen area, so that they can directly see the size of tumor regions. Fig. 6 illustrates this visualization method on an example WSI from the TCGA breast dataset. In the example shown in Fig. 6, the classes are normalized per sub-regions for each layer. In this example, the source material (organ) is 100% C50 (breast). This layer can be used to split slides with multiple samples to separate slides, similar to Plass et al. [56]. Layer 2 contains several tissue types. The most common tissue type in the example WSI is Connective-Tissue, which is further divided into Connective-Tissue-Fat (56.93%), Connective-Tissue-Dense (20.43%), Connective-Tissue-Loose (5.93%), Connective-Tissue-Blood (2.7%) and Connective-Tissue-Lymphatic (0.98%). Also, Epithelium with 8.16% and Nerve tissue with 0.28% are present. The main part in Layer 3 is Neoplastic-Malignant with more than 60%, followed by Not Encoded with almost 30%. Also, Neoplastic (5.91%) and Inflammation-Chronic (3.31%) alterations are present.

Area ratios as meta-information of tissue maps advance the search and filter capabilities of WSI archives. The visualization of the area ratios can be extended to multiple slides as shown in Fig. 7. Such a visualization in WSI archives allows to inspect multiple slides at the same time and extract relevant information from WSIs very quickly. Each WSI is represented with three bars. The first bar shows layer 1, the source material. One slide can contain several source materials. Typically, this occurs with biopsies of closely situated organs such as the colon and liver. The second and third bars show the area ratios of layer 2 (tissue types) and layer 3 (pathological alterations) respectively. This attention-based approach enables pathologists to more quickly select relevant WSIs for research purposes, even without directly inspecting them. Additionally, the proposed standard supports advanced filtering in WSI archives, allowing users to display only WSIs containing the desired content. This simplifies the work in pathological laboratories, and prospective pathologists can be trained more efficiently by a targeted selection of WSIs with specific content. In addition, the proposed standard provides AI-ready metadata and supports the creation of balanced and high-quality datasets, which are essential for reliable AI training and validation.

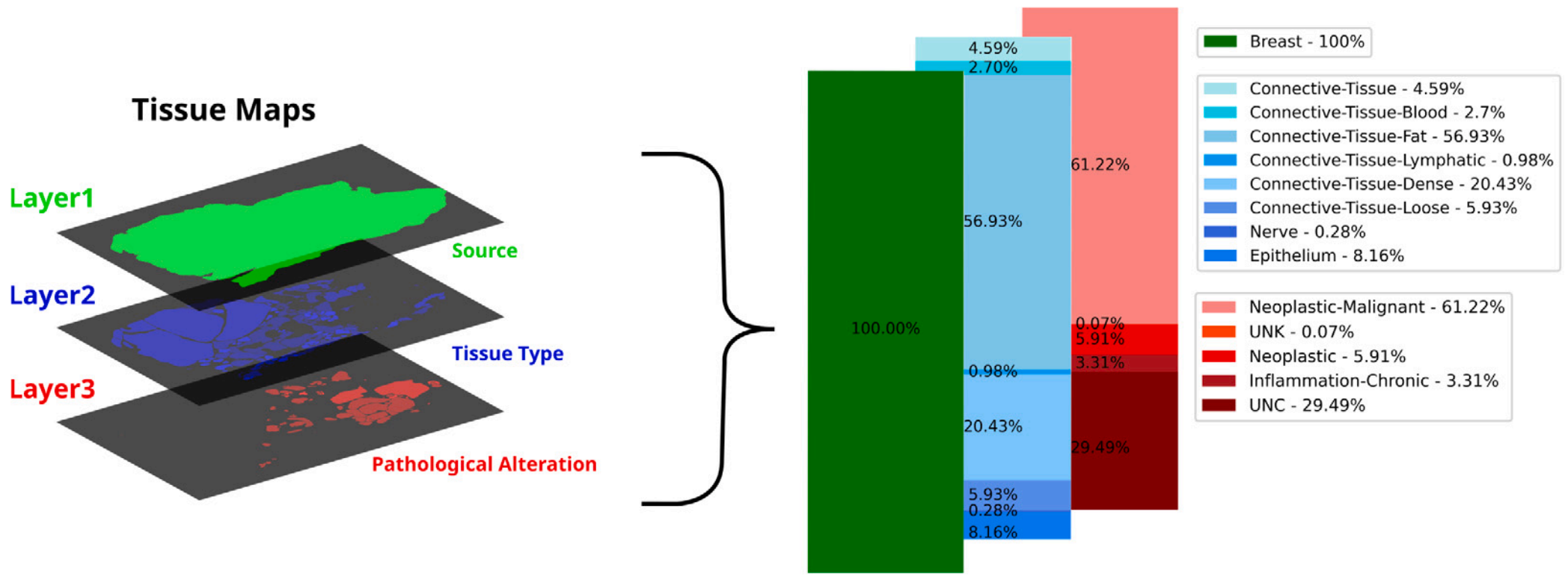

**Fig. 6.** Visualization of tissue composition from the example WSI. For better visibility, the areas are normalized per content of each tissue map layer.

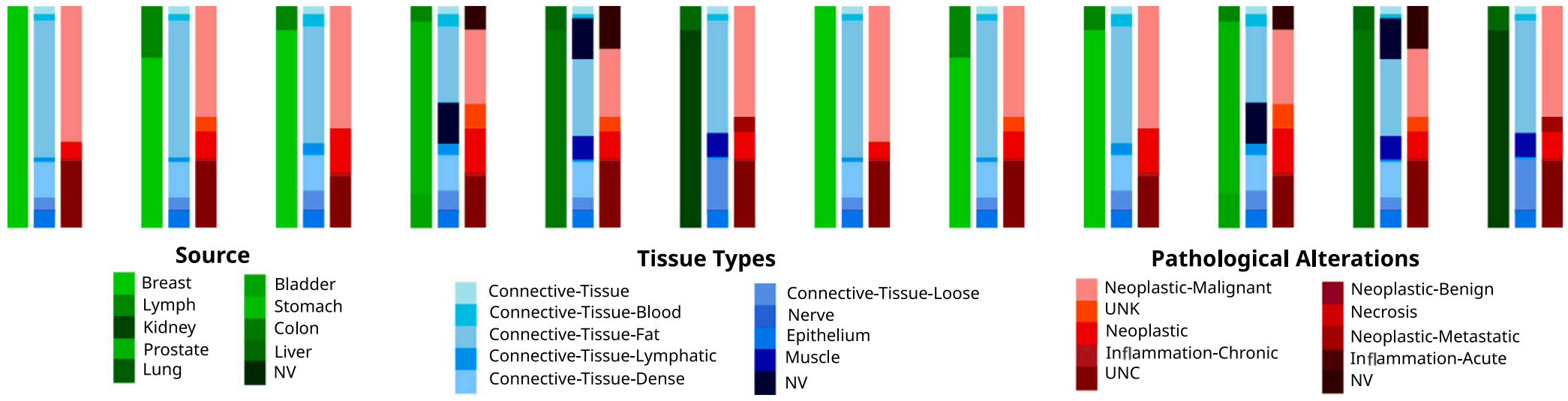

**Fig. 7.** Example of a possible visualization of WSI content in catalogs or WSI archives. Each WSI composition is represented with the three layers source material, tissue type and pathological alteration. The visualization color is defined by a LUT in the layer profiles.

## 4. AI-based metadata extraction for tissue maps

### 4.1. Dataset

A catalog was built with WSIs of 99 slides from TCGA dataset [1] across different organs like uterine, colon, lung, kidney, skin and more. The TCGA dataset contains regions of different tissue types and pathological alterations.

All slides were loaded into V7 Darwin [51] and only some areas of the WSIs were annotated based on the defined classes from the layer profiles of the tissue maps. Slides contain one or multiple annotation classes from tissue types and pathological alterations. Pathologists reviewed the annotations. The annotations of the TCGA slides are provided in the Github repository.

### 4.2. Tissue classifier for metadata extraction

Experiments are done with the integration of the proposed standard into WSI catalogs and for training AI algorithms. The overall design of the experiment was to obtain preliminary results by applying binary classifiers for defined tissue types and pathological alterations. Binary classifiers were used to estimate and limit the difficulty of the task. The metadata is extracted using these binary classifiers and then integrated into the test catalog as AI-ready metadata, according to the proposed standard.

Binary classifiers were used for 5 different tissue type classes, Connective-Tissue-Fat, Connective-Tissue, Muscle, Epithelium and Nerve and 6 pathological alterations Neoplastic, Neoplastic-Metastatic, Neoplastic-Malignant, Neoplastic-Benign, Necrosis, and Inflammation-Chronic, across different organs. Because of the partly annotated areas of the WSIs and the resulting unbalanced data, the pathological alterations were combined into one class and trained with a binary classifier.

From the annotated regions of each WSI, patches of size 512 × 512 pixels are extracted at pyramid level 1, which corresponds to an effective magnification of 10×. Patches are shifted with a stride of 128 pixels in $x$ and $y$ directions so that they overlap during the tiling process. The extracted patches are used for the training of the binary classifiers. Some patches have overlapping annotations. Hence, there are two approaches to managing overlapping annotations. First, the patch belongs to the tissue type if the tissue type is in the list of covered tissue types. The other method would only consider the patch belonging to a tissue type if the tissue type has the highest occurrence (strict labeling). In this work, strict labeling is used.

As a base model, VGG16 [57] was used as a feature extractor with pre-trained weights on ImageNet [58]. The model was modified to handle an input size of 512 × 512 pixels and fine-tuned for the pathology tasks. For a binary classification, a global max pooling layer and a classifier head consisting of a dropout layer were added followed by a fully connected layer. Each model was trained with the same settings: batch size of 32, learning rate of 5e−05, L2 regularization of 5e−05 and trained for a maximum of 50 epochs.

The binary classifiers are trained with 72 WSIs of the TCGA dataset and evaluated on 27 TCGA test slides. In total 242,716 patches are extracted from the annotated regions and used for training. Each classifier is evaluated on 27 partly annotated test WSIs. For the evaluation 172,378 patches are extracted from the annotated areas of the test slides and classified. The results are compared with the annotation classes and the average accuracy over the test WSIs for each classifier is calculated. The distribution of the extracted patches for training and testing is shown in Table 3. The patches are unbalanced, which can lead to biased multi-class classifiers. Therefore, a metadata description of the WSI content is essential to build balanced datasets for digital pathology. The proposed standard facilitates balanced sampling for AI training and validation, helping to reduce biases.

## 5. Experimental evaluation of the metadata model

The trained and validated classifiers are used to segment the entire WSIs of the test set. Each WSI is divided into patches. Each patch is classified and overlapping patches are averaged to generate a prediction segmentation map for each binary classifier. The segmentation maps of each classifier are combined to form the tissue maps of tissue type and pathological alterations. The tissue maps are combined with certain rules: Overlapping regions between neighboring patches are averaged, which can result in border artifacts. If more classifiers give a stronger prediction (above 50%), the area is marked as uncertain; if all classifier predictions are below 50%, the area is left unclassified otherwise, strict labeling is used. For pathological alterations, a segmentation map is created, where overlapping classified patches can result in different values (lower and above 50%). Such areas are labeled as uncertain. If overlapping patches are classified with stronger prediction (above 50%), the areas are labeled as pathological alterations. If the predictions are lower than 50% the area is labeled as unclassified.

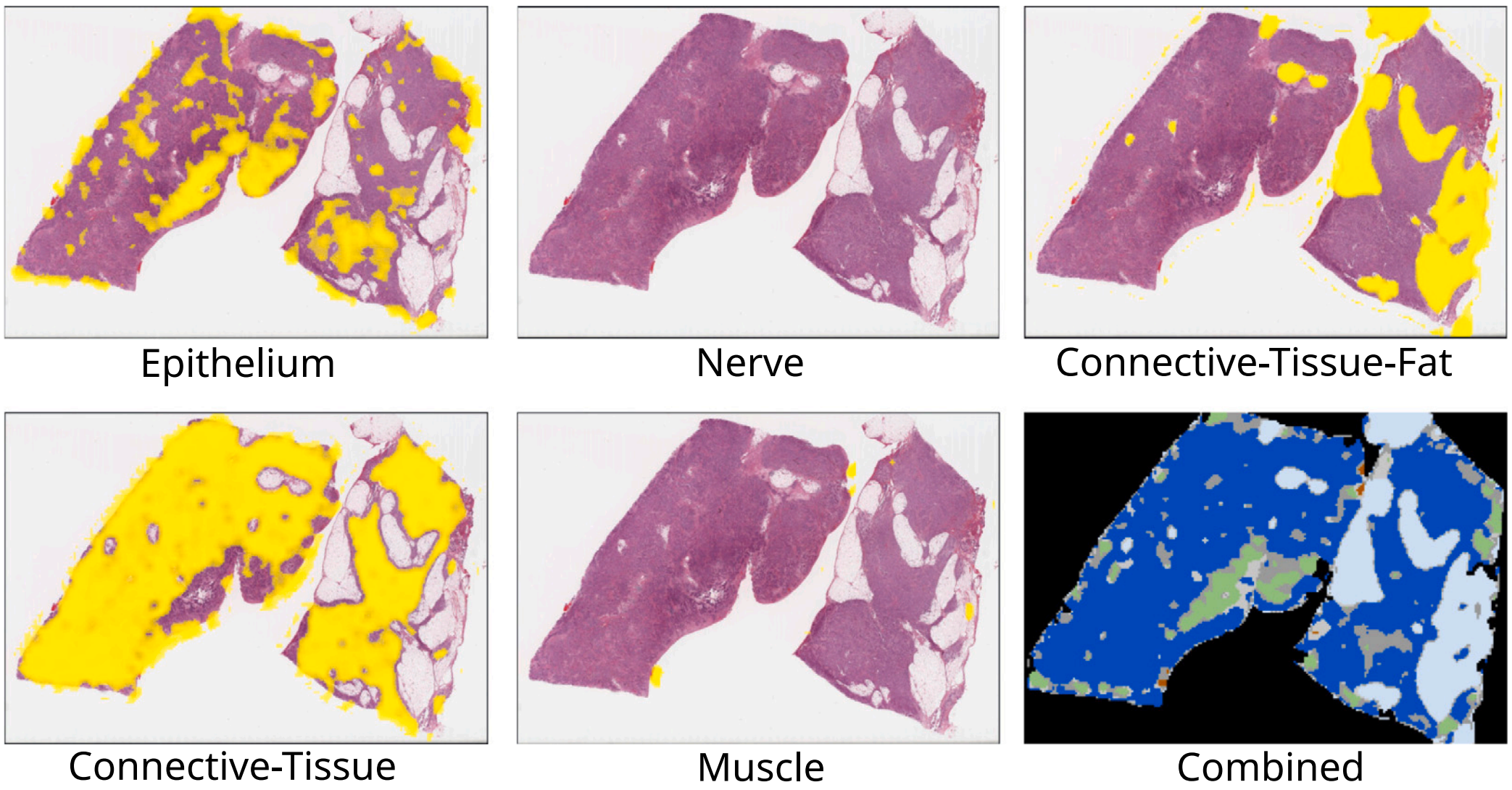

**Fig. 8.** Example segmentation maps for tissue type classifiers and the combined segmentation map with background (black), unclassified (light gray), uncertain (gray) and tissue regions with Epithelium (green), Nerve (yellow), Connective-Tissue-Fat (light blue), Connective-Tissue (blue) and Muscle (brown). (For interpretation of the references to color in this figure legend, the reader is referred to the web version of this article.)

**Table 3**
Number of patches extracted from the annotated regions for training and testing the binary classifiers. Binary classifiers are trained for each tissue type and one binary classifier is trained for all pathological alterations.

| Classifier | Train # of patches | Test # of patches |
|---|---|---|
| Tissue Types | | |
| Connective-Tissue-Fat | 31,706 | 16,984 |
| Connective-Tissue | 17,432 | 7,525 |
| Muscle | 9883 | 11,654 |
| Epithelium | 4210 | 2,215 |
| Nerve | 370 | 620 |
| Pathological Alterations | | |
| Neoplastic-Malignant | 144,715 | 99,689 |
| Neoplastic-Metastatic | 6342 | 14,555 |
| Neoplastic-Benign | 1031 | 834 |
| Necrosis | 21,748 | 14,139 |
| Inflammation-Chronic | 5288 | 4,163 |
| Pathological Alterations (total) | 179,124 | 133,380 |
| Total | 242,716 | 172,378 |

**Table 4**
Averaged accuracy of each binary classifier.

| Task | Accuracy |
|---|---|
| Connective-Tissue-Fat | 0.9476 |
| Connective-Tissue | 0.8972 |
| Muscle | 0.9209 |
| Epithelium | 0.9365 |
| Nerve | 0.9440 |
| Pathological Alterations | 0.9359 |

The evaluation of the classifiers is done with patches only from the partly annotated regions of the TCGA test slides. Patches are extracted from the annotated areas and each binary classifier is evaluated. Some class annotations are not present in most of the slides. E.g. Nerve tissue is only annotated in a few of the 27 TCGA test slides. Therefore, only the average accuracy over all test data for each binary classifier is presented in Table 4. There are 5 binary classifiers for tissue types and one combined binary classifier for pathological alterations. The overall accuracy of each binary classifier is very high. The entire WSIs were not evaluated and border cases might be missed due to the evaluation with only annotated areas.

### 5.1. Visualization of extracted metadata

Each classifier creates a prediction segmentation map for each tissue type and pathological alteration. The results are combined based on the classified patches with the defined classes inside the tissue maps. Each tissue map is independent of each other and separately analyzed. This results for tissue types in areas of uncertainty, marked as gray or unclassified (light gray). In the case of tissue type, light blue shows Connective-Tissue-Fat, blue shows Connective-Tissue, green highlights Epithelium, brown shows Muscle and yellow marks Nerve tissue, as shown in Fig. 8. Black shows the background.

The same approach is used for the pathological alterations, where pathological alterations are red, uncertain regions are gray, due to overlapping patches with different classifications, unclassified areas are light gray and the background is black. Results for both layers (left tissue types and right pathological alterations) are shown in Fig. 9. The results are evaluated with pathologists. The metadata for layer 1 (source) is extracted from the WSI description, and no classifier is required.

It can be seen, that the binary classifiers struggle with exact border delineation between different tissue types and pathological alterations. This is due to the patch-based approach for classification, strict labeling and the combination of each single prediction segmentation map to one combined segmentation map.

The combined segmentation maps for tissue types and pathological alterations are analyzed and the area ratios of the different classified areas are calculated. Fig. 10 shows the composition of the specimen with the predicted tissue types on examples of the test slides. The area ratios are normalized per specimen area (background excluded). Pathologists prefer this normalization. Meta information is used to define the organ type. In all 4 example WSIs, high densities of Connective-Tissue can be observed. The density of other tissue types varies. Each of the slides has a high density of pathological alterations.

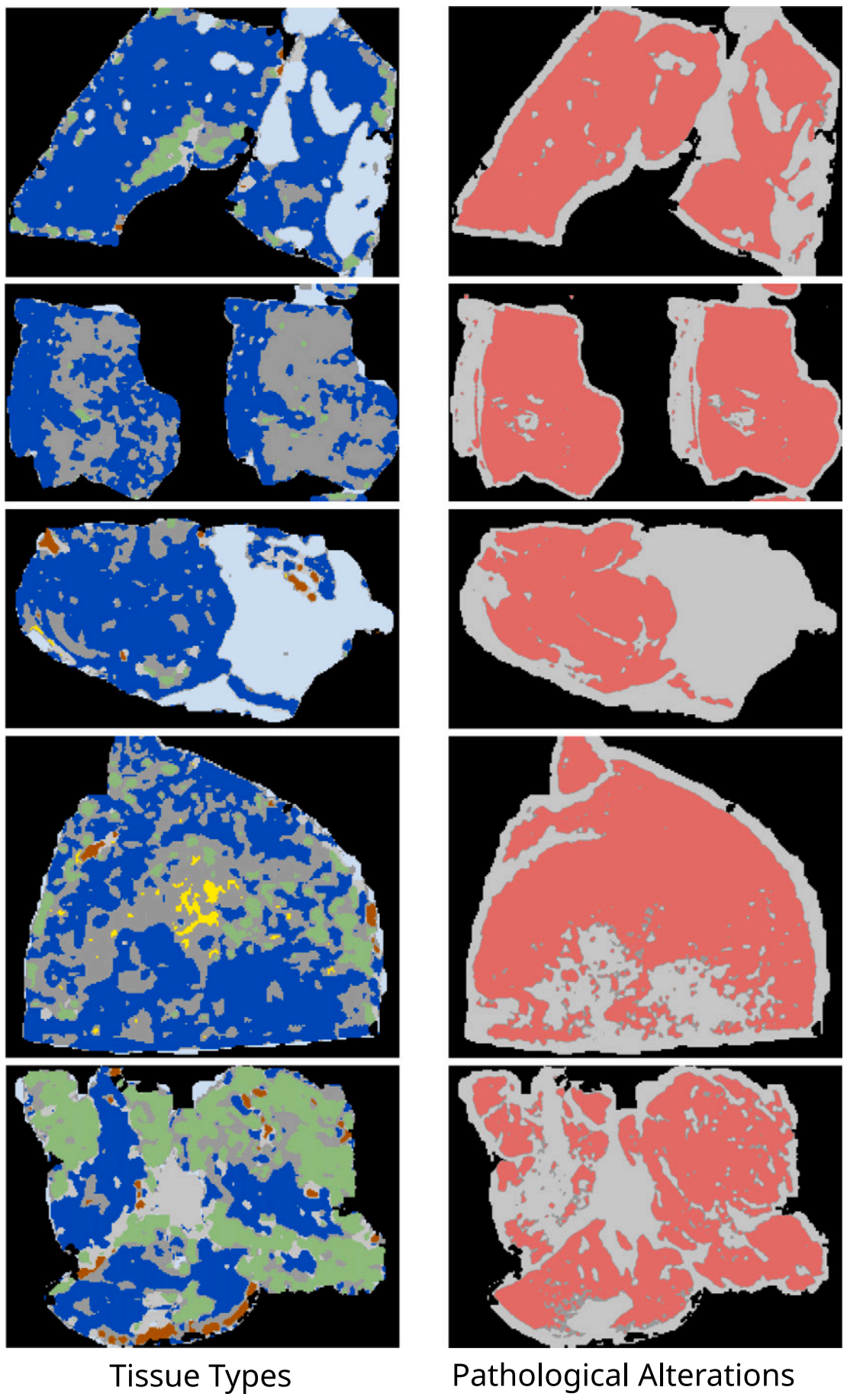

**Fig. 9.** Tissue maps of layers tissue type and pathological alteration visualized as segmentation maps for different WSIs of the test dataset. Tissue types of WSIs are shown on the left with background (black), unclassified (light gray), uncertain (gray), Epithelium (green), Nerve (yellow), Connective-Tissue-Fat (light blue), Connective-Tissue (blue) and Muscle (brown). The pathological alterations for the same WSIs are shown on the right with background (black), unclassified (light gray), uncertain (gray) and pathological alterations with Neoplastic, Neoplastic-Metastatic, Neoplastic-Malignant, Neoplastic-Benign, Necrosis, and Inflammation-chronic (red). Each layer is independent of each other. (For interpretation of the references to color in this figure legend, the reader is referred to the web version of this article.)

A full overview of the composition of the test dataset (WSI catalog) is shown in Fig. 11. The majority of the test slides contain a high density of Connective-Tissue with up to 77%. There are also slides with a high density of Muscle tissue and Epithelium. Some regions are unclassified and uncertain due to the small number of defined tissue classes. Nerve tissue is only present in small proportions in some of the slides. Most slides have a high concentration of pathological alterations around 50% and above. Such an analysis provides a detailed metadata description of the composition of individual slides and entire WSI cohorts, generating AI-ready metadata from tissues for AI training and validation. These metadata update databases with search parameters that filter WSIs based on specific composition criteria, facilitating interoperability between catalogs and research institutions, as well as the selection of slides and the creation of high-quality, balanced datasets for AI development and research.

## 6. Discussion

This paper introduces a cross-catalog metadata standard with multi-layer tissue maps to describe the morphology of WSIs for large-scale WSI archives. It is based on a three-layer model with source, tissue type and pathological alterations. Each layer forms a tissue map based on a defined profile with common syntax and notation to fulfill the requirements of surgical pathology or biology use cases. Integrating tissue maps into WSI archives enhances interoperability between catalogs and research institutions, enables faster and more efficient WSI selection, and supports the creation of balanced, high-quality datasets. Moreover, the tissue maps produce AI-ready metadata utilized for both AI training and testing.

Furthermore, a framework WSIDOM is provided, to visualize the tissue maps as segmentation maps by using LUT defined in the layer profiles of the metadata standard. This allows human readability of the tissue map layers.

Currently, vision transformer [59] based on self-supervised learning [60] or multiple instance learning [14] are popular for digital pathology: Transformer-based models [22,61,62] are used to predict survival rates [21,22,61], classify/detect cancer types [61,63,64], and also for tissue segmentation [20,65].

But ML models do not yet process entire WSIs at once due to the WSI's big resolution of up to $150,000 \times 150,000$ pixels. Therefore, WSIs are divided into small patches and at each training iteration a subset of these patches is sampled [22]. Researchers are trying to increase the size of the patches [61,62] across different pyramid levels. The size of a patch can be increased up to $32,768 \times 32,768$ pixels [64] by applying dilated attention [66] and with an up-scale framework TORCHSCALE [67]. Larger patch sizes require more advanced hardware platforms, which are often not available.

However, random sampling may be suboptimal, particularly for unbalanced datasets. The probability of sampling a random patch in an image with a specific pathological alteration is determined by the specific pathological alteration and the full image area. The same applies to tissue types and organ types. Random sampling can lead to underrepresented classes and imbalanced training data, which consequently result in biased models. That is why binary classifiers were trained for each of the tissue type classes and pathological alterations to generate segmentation maps, which were combined to tissue maps. Patch extraction for training multi-classifiers benefits from the 8-bit encoding that represents the defined classes of the tissue maps. With a more advanced WSI selection process, slides can be chosen based on fine-grained information about the specimen composition. This helps to mitigate data imbalance in the training of ML models by enabling uniform sampling of training data across classes.

Another use of the tissue maps for ML in digital pathology involves utilizing the class assignments from the tissue maps as feature representation for ML, as shown in Fig. 12(a). The class assignments from the tissue maps are concatenated with the raw image data of the WSI. Each of the layers corresponds to a dimension in this novel feature space. By providing this contextual information together with raw pixel values, ML models get a much deeper understanding of the WSI content. Furthermore, the three layers allow a representation as a single RGB color image, which is suitable as input features for the development of ML algorithms in digital pathology [13,15]. Tissue maps enable a more effective and efficient learning of the models and help to counteract unbalanced training data.

This paper introduces the first experiments with binary classifiers for different tissue types and pathological alterations to automate the process of tissue map generation from WSIs and the metadata analysis about the content of WSIs across catalogs. It shows on the example of a TCGA dataset the composition of each WSI in the test set. It demonstrates the advantages of the defined layer profiles and the generated metadata of each tissue map for WSI search and dataset creation. This paper demonstrates how WSIs can be represented as multi-layer tissue maps, serving as a metadata standard for various applications. Additionally, foundation models can be used to classify [16,35,68], segment [17,68] or annotate [19] histological images or predict pathological workflow parameters [34].

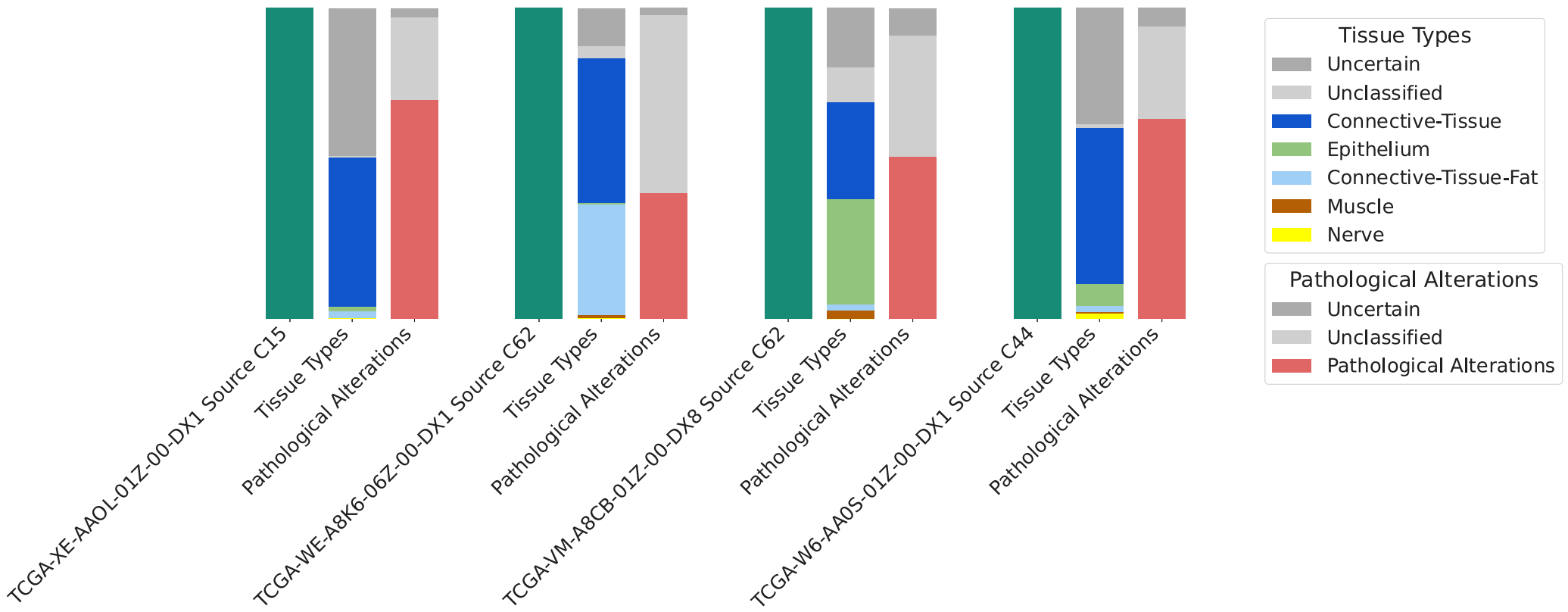

**Fig. 10.** Composition analysis of the layers source (organ), tissue types and pathological alterations for example WSIs of the test dataset. The first bar shows the organ source material, the second shows the composition of the tissue type layer and the third bar shows the composition of the pathological alteration layer for different WSIs.

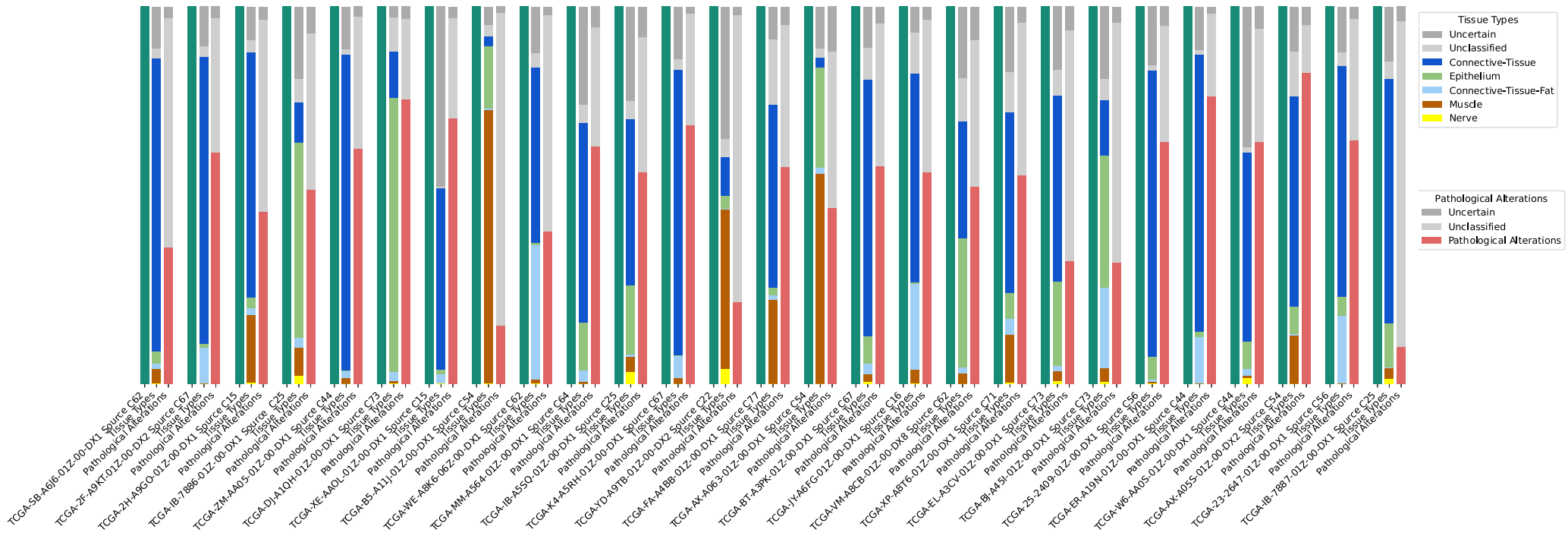

**Fig. 11.** Composition analysis of the layers source (organ), tissue types and pathological alterations for TCGA test slides. The composition of each WSI is represented with area ratios of the specimen with bars for source material, tissue type and pathological alterations.

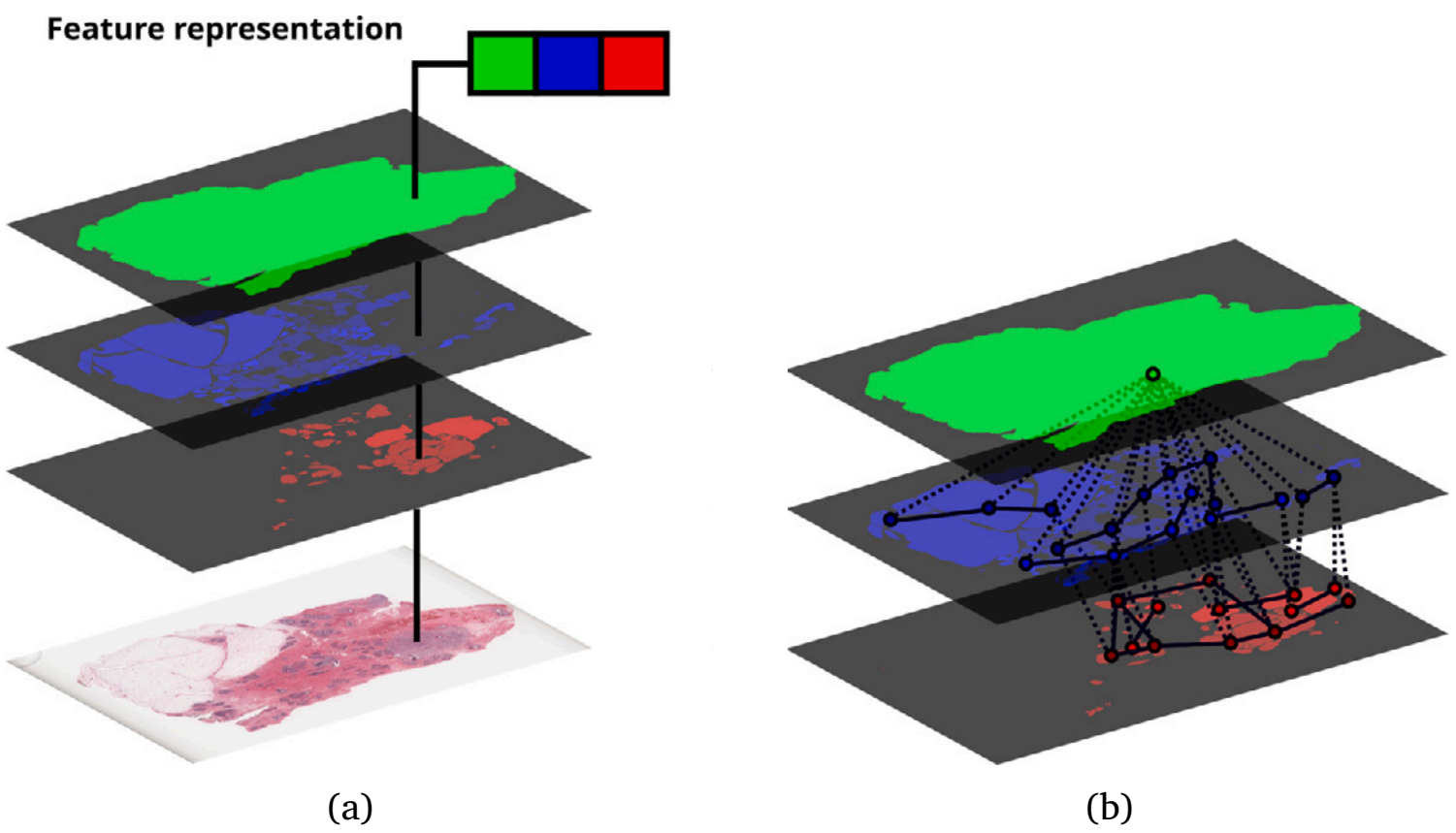

**Fig. 12.** Examples of possible WSI representations. (a) Feature representation with the tissue maps and their associated color channels of an RGB image. (b) Graph-based WSI representation by using the tissue maps.

However, the metadata model approach presented in this paper also includes the WSI metadata directly into WSI archives as multi-layer tissue maps, where this information enables filter capability in biobank management tools such as OpenSpecimen [69]. This gives a big advantage for the selection of WSIs for future research projects. Additionally, generated bar charts can be leveraged for similarity search similar to approaches such as SMILY [31] or Yottixel [32]. In contrast to patch-based methods, this approach relies solely on extracted metadata to construct the bar-chart representations, eliminating the need for WSI patch extraction. This substantially reduces computational overhead and enables more efficient large-scale similarity search.

**Table 5**
Comparison of digital pathology tools and metadata models, highlighting their search motivation and corresponding metadata granularity.

| Tool | Search paradigm | Metadata structure | Granularity |
|---|---|---|---|
| Metadata model | Content-based, tissue map, metadata-driven | tissue map, hierarchical annotations | Slide, source, tissue types, pathological alterations |
| Yottixel [32] | Content-based, image similarity | Patch-based embeddings | Slide, region |
| SMILY [31] | Content-based, image similarity | Patch-based embeddings | Slide, region |
| SISH [33] | Content-based, hybrid | Patch + slide-level info | Patch, slide |
| MIKAIA [17] | AI-assisted analysis | Segmentation map, hierarchical annotations | Patch, tissue type |

In recent years, researchers also try to extract spatial information with graph transformers from different magnification levels of WSIs and learn the relationships of these levels to build classifiers or detectors for different pathological tasks [70–73]. Each of these works takes patches from different magnification levels of the WSI and builds frameworks to construct graphs based on the information of these patches to improve different classification tasks.

With tissue maps a spatial relationship between the 3 layers (source material, tissue types, pathological alterations) is established by forming a graph-based representation with pathological entities of the WSI [70,72]. In this graph-based representation, the nodes of the graph represent the classes of segmented areas of the different layers and the weights represent the distance of the spatial relationship between the nodes. Such a graph-based representation of a WSI is shown schematically in Fig. 12(b). Providing tissue maps with the three layers source material, tissue types and (pathological) alterations, enables training of ML models with graph-based methods [70–73]. The models benefit from the spatial relationships between each of these layers to find dependencies related to cancer research.

The three layers of the tissue map enable the extraction of additional information from WSIs, such as WSI content-specific area ratios for present source material types (organs), tissue types or (pathological) alterations. WSI archives include the WSI-specific information as metadata.

A test dataset is used to show the integration of the proposed metadata standard into WSI archives, which allows to build large-scale WSI archives with specific search capabilities from organs/organ types, tissue types, pathological alterations to specimen composition of WSIs across catalogs. WSI archives include the WSI-specific information as metadata. By providing detailed information about WSI content, the search functionality of WSI archives is greatly enhanced, allowing WSIs to be filtered according to the specific parameters required by pathologists or researchers. Such particular parameters can be the source material, tissue types or (pathological) alterations represented by the three layers of the tissue map. The calculation of the area for each defined entry in each layer allows the selection of desired WSIs of the WSI archive to build WSI catalogs based on these search parameters. The search capability will be included in biobank management systems to enable new possibilities for faster and more precise WSI selections for different research fields. This enables the construction of specific and balanced WSI catalogs for medical studies and the improvement of AI algorithm training and validation [74,75] in digital pathology, as datasets for these purposes are created and selected from archives more efficiently and effectively based on the enhanced search capabilities. A comparison of different tools with the metadata model is shown in Table 5 highlighting the differences of the search paradigm, metadata structure and search granularity. The metadata model is a different approach compared to the image query-based approaches described by Hedge et al. [31], Kalra et al. [13,32] and Chen et al. [33], where query images are used to find images with similar morphological structures, such as histopathological features, cancer grading or cancer subtyping. MIKAIA [17] is an AI-assisted analysis tool and is limited to specific tissue types and does not include a search functionality for WSI archives. However, this tool can be used for tissue map generation.

Incorporating prior knowledge of the content within a WSI holds substantial potential to enhance the performance of AI algorithms during both the training and inference phases. Standardized interfaces foster seamless integration, comparability, and collaboration across diverse platforms and stakeholders. In this context, the EcosysteM for Pathology Diagnostics with AI Assistance (EMPAIA) [74,75] has been pivotal in developing APIs that support such connectivity and interoperability. EMPAIA serves as a comprehensive ecosystem for digital pathology, promoting collaboration among pathologists, industry leaders, and computer scientists. It provides a robust framework for evaluating various AI algorithms, thereby streamlining the adoption and assessment of new technological innovations in the field.

EMPAIA's suite of ML algorithms supports pathological diagnostics by offering a structured approach to testing and incorporating diverse analytical tools. The planned integration of the multi-layer tissue maps within EMPAIA's platform represents a significant advancement, enabling a more refined evaluation of digital pathology solutions. Embedding the proposed metadata standard through a standardized interface will enhance the process of selecting WSIs based on tissue maps and defined criteria, making the training and assessment of algorithms more efficient. Through this integration, EMPAIA will strengthen its capability to support precise and efficient algorithm evaluation and deployment in digital pathology.

## 7. Conclusion

This paper introduces a metadata model framework to generate 2D index maps, also called tissue map, to profile the content of WSIs. The tissue map is organized into three layers with source material (organ), common tissue types and (pathological) alterations providing fine-grained information about the content of the WSI. Segments of the WSI are assigned to predefined classes within these layers. This layered approach enhances WSI collections by supporting more effective and efficient content-based search.

An experimental evaluation in the pathology domain demonstrates the interoperability of the proposed metadata model by employing standardized syntax and semantics across different catalogs. Binary classifiers are used to automatically extract WSI content from the slides and generate tissue maps according to the proposed standard. The tissue maps are integrated into a test catalog based on TCGA WSIs. This test catalog illustrates the metadata model's utility and its integration into a WSI archive with a focus on improved search functionality and providing AI-ready metadata.

Beyond improving search, this framework unlocks new opportunities for AI development. By enabling faster creation of balanced, high-quality datasets, it supports more reliable training and validation of AI models. The AI-ready metadata generated through tissue maps reduces biases in training, while the structured multilayer design establishes meaningful relations between source material, tissue types, and pathological alterations. Together, these contributions provide a foundation for advancing AI in digital pathology by ensuring both robust dataset assembly and transparent, interpretable metadata.

An open-source framework, WSIDOM, is developed to support visualization and conversion of annotations from commonly used tools into tissue maps. WSIDOM is user-friendly, adaptable to different tissue map profiles, and designed to meet the needs of diverse research applications. The framework is available at: https://github.com/human-centered-ai-lab/WSIDOM.

Future work will focus on leveraging a larger and more balanced set of annotations to train a multi-class tissue segmentation model at patch level. This will further automate metadata extraction based on predefined tissue map profiles. Extending this approach to various WSI catalogs will facilitate advanced, content-aware search capabilities across WSI repositories. In addition, similarity search approaches based on generated bar charts derived from the extracted metadata should be implemented. Such approaches would allow efficient retrieval of similar slides without requiring patch-level extraction, relying solely on the summarized metadata, thereby substantially improving search performance and scalability across large WSI collections.

## CRediT authorship contribution statement

**Gernot Fiala:** Writing – review & editing, Writing – original draft, Visualization, Validation, Software, Methodology, Investigation, Formal analysis, Data curation, Conceptualization. **Markus Plass:** Writing – review & editing, Writing – original draft, Visualization, Validation, Supervision, Resources, Methodology, Funding acquisition, Conceptualization. **Robert Harb:** Writing – review & editing, Software, Formal analysis, Conceptualization. **Peter Regitnig:** Writing – review & editing, Validation, Data curation, Conceptualization. **Kristijan Skok:** Writing – review & editing, Validation, Data curation. **Wael Al Zoughbi:** Writing – review & editing, Validation, Data curation. **Carmen Zerner:** Writing – review & editing, Data curation. **Paul Torke:** Writing – review & editing, Investigation, Data curation, Conceptualization. **Michaela Kargl:** Writing – review & editing, Writing – original draft, Methodology, Conceptualization. **Heimo Müller:** Writing – review & editing, Supervision, Methodology, Funding acquisition, Conceptualization. **Tomas Brazdil:** Writing – review & editing, Software. **Matej Gallo:** Writing – review & editing, Software. **Jaroslav Kubín:** Writing – review & editing, Software. **Roman Stoklasa:** Writing – review & editing, Software, Project administration. **Rudolf Nenutil:** Writing – review & editing, Validation, Data curation. **Norman Zerbe:** Writing – review & editing. **Andreas Holzinger:** Writing – review & editing. **Petr Holub:** Writing – review & editing, Conceptualization.

## Funding

Parts of this work have received funding from the Austrian Science Fund (FWF), Austria, Project P-32554 (Explainable Artificial Intelligence) and the European Union's Horizon Europe research and innovation program under No. 101097036 (ONCOSCREEN). This publication reflects only the authors' view and the European Commission is not responsible for any use that may be made of the information it contains.

## Declaration of competing interest

The authors declare that they have no known competing financial interests or personal relationships that could have appeared to influence the work reported in this paper.

## Acknowledgments

We thank our colleagues at BBMRI-ERIC and BBMRI.at for their valuable advice regarding the annotation classes, and the IT department of the Medical University of Graz for hosting the annotation infrastructure. We are also grateful to Bettina Monschein for proofreading support and to the ONCOSCREEN consortium for their collaboration. Parts of the work were done with data generated by the TCGA Research Network: https://www.cancer.gov/tcga.

## Data sharing statement

We used the public datasets from The Cancer Genome Atlas (TCGA) (https://www.cancer.gov/ccg/research/genome-sequencing/tcga).